\documentclass[preprint,12pt]{elsarticle}

\usepackage[printwatermark]{xwatermark}
\usepackage{xcolor}
\usepackage{graphicx}
\usepackage{lipsum}

\usepackage{graphicx}
\usepackage{amssymb}

\usepackage{lineno}

\journal{Cognitive Systems Research}

\begin{document}

\begin{frontmatter}


\title{Populations of Spiking Neurons for Reservoir Computing: Closed Loop Control of a Compliant Quadruped.}



\author{Alexander Vandesompele, Gabriel Urbain,
Francis wyffels, Joni Dambre}

\address{IDLab-AIRO, Electronics and Information Systems Department, Ghent University - Imec, Ghent, Belgium.\\
\textcopyright 2019, Elsevier. Licensed under the Creative Commons Attribution-NonCommercial-NoDerivatives 4.0 International http://creativecommons.org/licenses/by-nc-nd/4.0 \\
DOI: 10.1016/j.cogsys.2019.08.002}

\begin{abstract}
Compliant robots can be more versatile than traditional robots, but their control is more complex. The dynamics of compliant bodies can however be turned into an advantage using the physical reservoir computing framework. By feeding sensor signals to the reservoir and extracting motor signals from the reservoir, closed loop robot control is possible. Here, we present a novel framework for implementing central pattern generators with spiking neural networks to obtain closed loop robot control. Using the FORCE learning paradigm, we train a reservoir of spiking neuron populations to act as a central pattern generator. We demonstrate the learning of predefined gait patterns, speed control and gait transition on a simulated model of a compliant quadrupedal robot.
\end{abstract}


\end{frontmatter}


\section{INTRODUCTION}
\label{S:1}

Compliant robots can provide a greater robustness, flexibility and safety compared to traditional, stiff robots \citep{pfeifer2007self}. However, the control paradigms used in traditional robotics cannot be applied to compliant robots, due to the complexity of predicting the state of the compliant body. The complex dynamics can however be turned into an advantage with the concept of embodied computation (also referred to as morphological computation), where the physical body is treated as a computational resource \citep{hauser2011towards,pfeifer2007self,fuchslin2013morphological}.\\

Physical reservoir computing provides a framework for harvesting the body as a computational resource~\citep{caluwaerts2013locomotion}. Monitoring the non-linear body dynamics of a compliant body can be a useful source of information. In some systems extremely little additional computation is required to accomplish a task, for instance locomotion control of a tensegrity robot~\citep{caluwaerts2013locomotion} and control of a soft robotic octopus arm~\citep{nakajima2013soft}. By combining the body feedback with some additional computational power (e.g. a 'brain'), more complex locomotion tasks can be accomplished~\citep{degrave2015developing}. The computations that naturally occur in the body are then augmented with a small 'brain' to achieve partially embodied control. In~\cite{degrave2015developing} this was found to be necessary for gait generation with a quadruped robot. In~\cite{burms2015reward} and~\cite{urbain2017morphological}, more complex tasks were addressed using, respectively, a tensegrity robot and a mass-spring network.\\

An example of a low level brain function is the generation of rythmic activity by central pattern generators (CPG). CPGs are neural networks in the spinal cord of vertebrate animals, that have been observed to generate rythmic activity and are involved in rythmic movements such as locomotion and respiration~\citep{delcomyn1980neural}. Even though biological CPGs can be active without sensory input or descending input from other brain regions, both inputs can modulate the CPG. In decerebrated cat experiments, gait frequency and even gait transition can be controlled with a simple electrical stimulus to the spinal cord~\citep{shik1966orlovskii}. Other decerebrated cat experiments revealed that also sensory inputs can modulate the ongoing rythmic activity (reviewed in~\cite{rossignol1993intralimb}).\\

CPGs can be implemented with a neural network by using the reservoir computing framework~\citep{wyffels2009design}. In reservoir computing, a reservoir is excited by inputs and provides a spatiotemporal expansion of this input. The reservoir is typically a randomly connected neural network, of which the weights are rescaled such that the network operates at 'the edge of chaos' \citep{legenstein2007edge}. The output is then a linear mapping of the reservoir activity. In our setup, by feeding body sensors of a robot to a randomly connected reservoir, a spatiotemporally enriched interpretation of the body sensors is created. Thanks to this expansion, more complex patterns can be extracted from the reservoir activity using only linear regression. Reservoir computing with spiking neurons is traditionally performed with a liquid state machine~\citep{maass2002real}. Here, we propose using population coding, where the unit of the reservoir is a population of spiking neurons. Whilst this method allows to apply the same principles as in the well established rate based reservoir computing, it also allows to potentially profit from using a spike based implementation. The number of tunable parameters, both at neuron and population level allows for optimizing reservoir dynamics for closed loop dynamical systems. Additionally, efficient hardware implementations (e.g. SpiNNaker, ~\cite{furber2014spinnaker}) could allow to run the network with low power usage on mobile robots. Lastly, this framework allows interfacing with spike-based sensors (e.g. the DVI camera, ~\cite{lichtsteiner2008128}) that provide low latency and low redundancy sensor data.\\

\begin{figure}[thpb]
      \centering
      \framebox{\parbox{3in}{\includegraphics[scale=0.128]{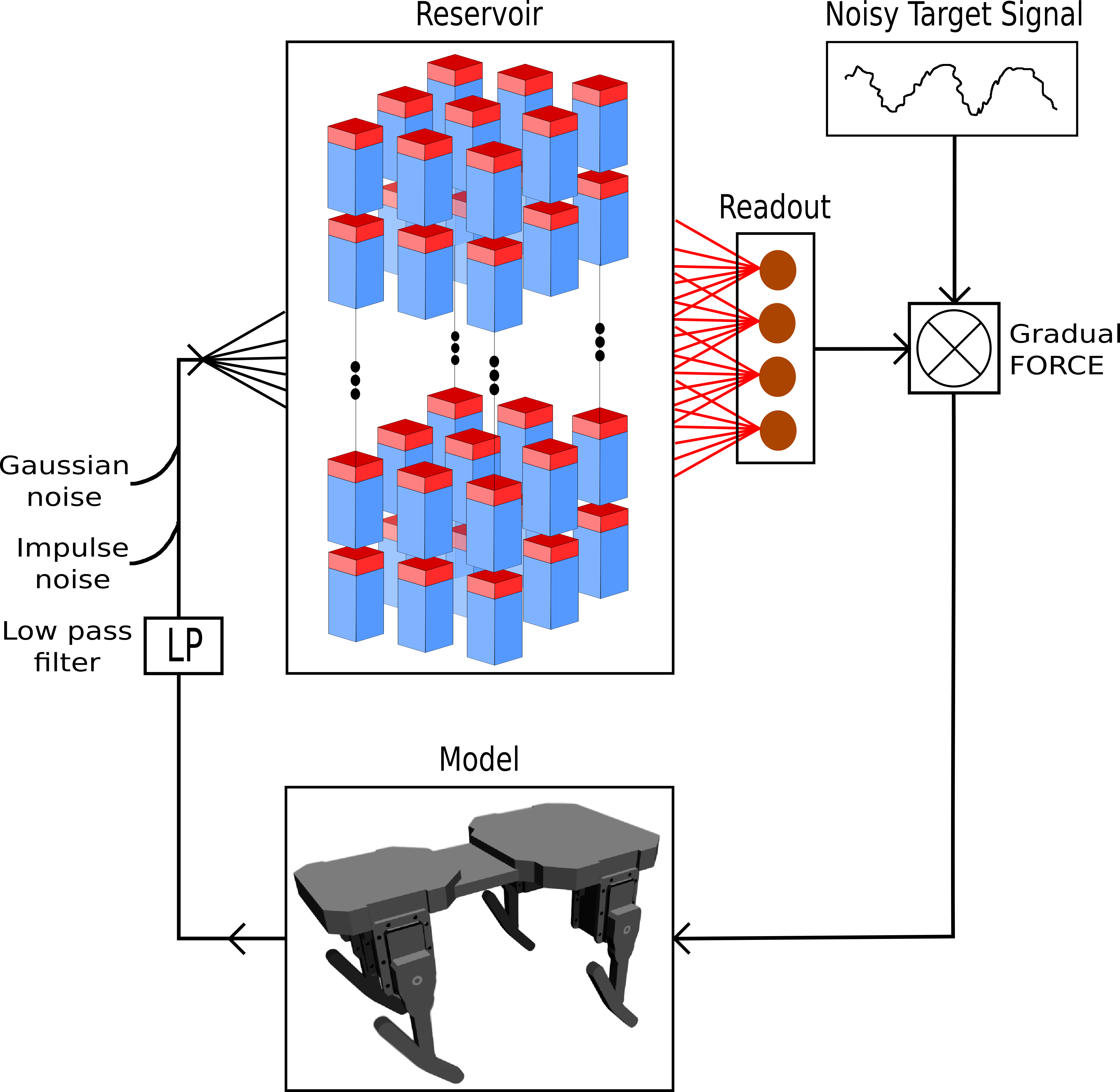}}}
      \caption{Overview of the closed loop control system. Learned connection weights in red.}
      \label{overview}
   \end{figure}

In this paper, we demonstrate the feasability of using populations of spiking neurons in embodied computation by creating stable closed loop locomotion control for a compliant robot. To achieve this, we applied the physical reservoir computing framework to a simulated model of the Tigrillo robot~\citep{willems2017quadruped}, a compliant quadrupedal platform. We add a 'brain' to the robot which is also a reservoir, consisting of spiking neurons. This neural network is trained to function as a CPG and, similar to biological CPGs, can be modulated by both body sensors and simple control inputs. To create a stable dynamical system, capable of generating robust periodic movements, online linear regression can be applied (FORCE learning,~\cite{sussillo2009generating}) in a gradual fashion~\citep{caluwaerts2013locomotion}. In \cite{nicola2017supervised}, FORCE learning was applied to spiking neural networks for the extraction of complex, dynamical signals. Here, we apply FORCE learning for closed-loop locomotion of a robot model, introducing the robot body in the loop. Figure~\ref{overview} presents an overview of the implemented system. Four readout neurons are trained to produce motor signals for the actuated joints of the Tigrillo model. Four body sensors, sensing the angle of the passive joints, are fed as input to the neural network.\\

In the next section the components of the closed loop system are presented and the learning algorithms involved are detailed, including a method for monitoring the reservoir state. The results section presents different gait patterns that have been learned, as well as gait frequency control and gait switching control.\\

\section{MATERIALS \& METHODS}

\subsection{Simulation}

\begin{figure}[thpb]
      \centering
      \framebox{\parbox{3.8in}{\includegraphics[scale=0.2]{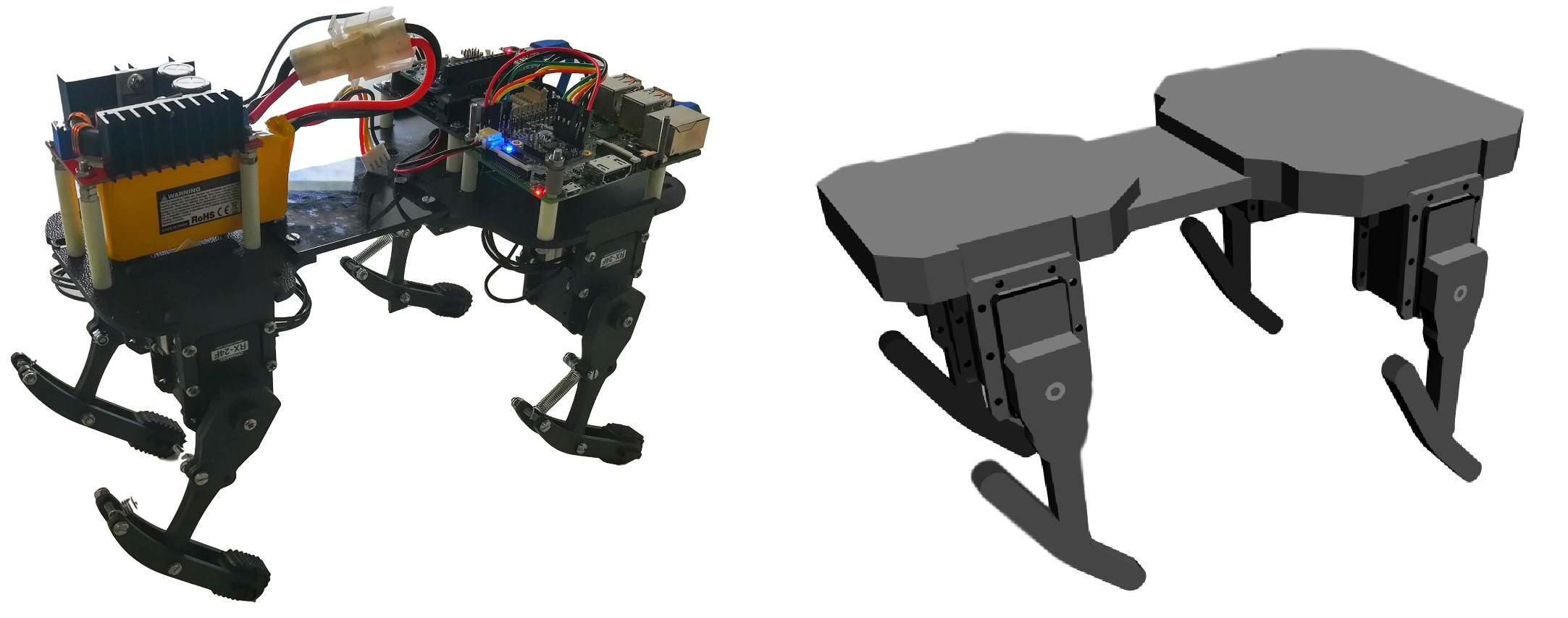}}}
      \caption{Tigrillo physical robot (left) and model (right).}
      \label{tigrillo}
   \end{figure}

The model used in this work is based on the physical Tigrillo robot~\citep{willems2017quadruped}, see Figure~\ref{tigrillo}. Tigrillo is a low-cost platform developed for researching compliance in quadrupeds. The robot has four legs, consisting of two joints: one actuated with a servo motor (hips and shoulders) and one passive joint (knees and elbows). The passive joints are loaded with a spring, providing compliance. The angle of the passive joints (on the physical robot measured with Hall effect sensors) reflects the state of the robot body and its interaction with its environment, and is therefore useful as a sensor input in the closed loop control system. The Tigrillo model is a parametrized stick and box model that mimicks the weight distribution and physics of the physical robot. \\

All simulations have been performed on the Neurorobotics Platform (NRP,~\cite{falotico2017connecting}). The NRP provides an interface between an environment simulator (Gazebo) and a spiking neural network simulator (e.g, NEST,~\cite{gewaltig2007nest}). In this work, ODE~\citep{drumwright2010extending} was used as physics engine.\\

\subsection{CMA-ES}\label{sectioncmaes}

To find hip joint motor signals for gaits that suit the body dynamics, the covariance matrix adaptation evolutionary strategy (CMA-ES) algorithm~\citep{hansen2001completely} is used. CMA-ES is an evolutionary algorithm that samples solutions from a multi-variate normal distribution. Every iteration, the mean and the covariance matrix of the distribution are updated. The mean is updated to increase the likelihood of previously successful solutions. The covariance matrix is updated to increase the likelihood of a previously successful search step. CMA-ES can handle non-convex fitness landscapes with many local maxima well. It requires few initial parameters and doesn’t require derivation of the search space.\\

The CMA-ES is used to optimize parameters of a parametrized CPG (described in~\cite{gay2013learning}). This CPG is implemented by a set of three equations:  
   
\begingroup
\large
    \begin{equation} \label{eq:cpg1}
        \begin{array}{rcl}
            \dot{r} &=& \gamma(\mu - r^2)r \\
            \dot{\phi} &=& \omega \\
            \lambda &=& r\, \textrm{cos}(\phi_L) + o \\
       \end{array}
    \end{equation}
\endgroup
   
Where $\phi$ and $r$ describe the current phase and the radius of the oscillator, respectively. Both are used to calculate the actual control value $\lambda$ (in degrees). $\mu$ is the target amplitude of the oscillator and $\gamma$ is  a  positive  gain that defines the speed of convergence of the  radius to the target amplitude. $o$ is the offset and $\omega$ the radial frequency of the oscillator. $\phi_L$ is a filter applied on the phase of the oscillator and is different for the stance and swing phase of the control as determined by the duty factor ($d$):

\begin{equation} \label{eq:cpg2}
    \begin{array}{rcl}
        \phi_L &=& \left\{
                    \begin{array}{ll}
                        \frac{\phi_{2\pi}}{2d} \qquad \qquad \qquad \qquad \quad\enskip  \textrm{if} \quad  \phi_{2\pi}<2\pi d\\
                        \\
                        \frac{\phi_{2\pi} + 2\pi (1-2d)}{2(1-d)} \quad \textrm{otherwise}\\
                    \end{array}
                  \right.\\\\
        \textrm{and} \; &\phi_{2\pi} = \phi \;(\textrm{mod} \; 2\pi)\\
    \end{array}
\end{equation}
   
The Tigrillo platform has four actuated hip joints controlled by four phase-coupled CPGs. The front left leg is chosen as reference leg and the phase difference of the remaining 3 legs is described by three phase offset ($po$) parameters. This is implemented by adding a term to the formula for the phase ($\phi$) in equation \ref{eq:cpg1}). For instance, for the coupling between the front left and front right oscillators:

\begingroup
\large
    \begin{equation} \label{eq:cpg3}
        \begin{array}{rcl}
            \dot{\phi}_{fr} &= \omega + w_{fr}sin(\phi_{fl}-\phi_{fr}-po_{fr})\\
         \end{array}
    \end{equation}
\endgroup

with $w_{fr}$ the coupling strength.\\

Initial parameters had a Gaussian distribution with 0.5 mean and 0.2 SD. Each generation consisted of 25 individuals, other parameters were kept at default as described in \cite{hansen2006cma}. Different gaits were found by optimizing different subsets of parameters. The parameters optimized in the search for the walking gait are listed in Table \ref{params}. The CPG frequency was kept constant at 1.44 Hz. The distance travelled from the origin was used as fitness function.\\

\begin{table}
\centering
\begin{tabular}{c c c c}
\\\hline
Parameter & Symbol & Range & Unit\\\hline
Front amplitude & $\mu_{f}$  & $[20,\:140]$ & degrees\\
Hind amplitude & $\mu_{h}$ & $[20,\:140]$ & degrees\\
Front duty cycle & $d_{f}$ & $[0.15,\:0.85]$ & \textit{NA}\\
Hind duty cycle & $d_{h}$ & $[0.15,\:0.85]$ & \textit{NA}\\
Front offset & $o_{f}$ & $[-60,\:60]$ & degrees \\
Hind offset & $o_{h}$ & $[-60,\:60]$ & degrees \\
Front right phase offset & $po_{fr}$ &  $[150,\:210]$ & degrees\\
Hind left phase offset & $po_{hl}$ & $[240,\:300]$ & degrees\\
Hind right phase offset & $po_{hr}$ & $[60,\:120]$ & degrees\\
\hline\\
\end{tabular}
\caption{\label{params}Parameters and their ranges included in the \textit{CMA-ES} optimization for the walking gait.}
\end{table}
   
\subsection{The Neural Network}
\label{brainsection}

\begin{figure}[thpb]
      \centering
      \includegraphics[scale=0.27]{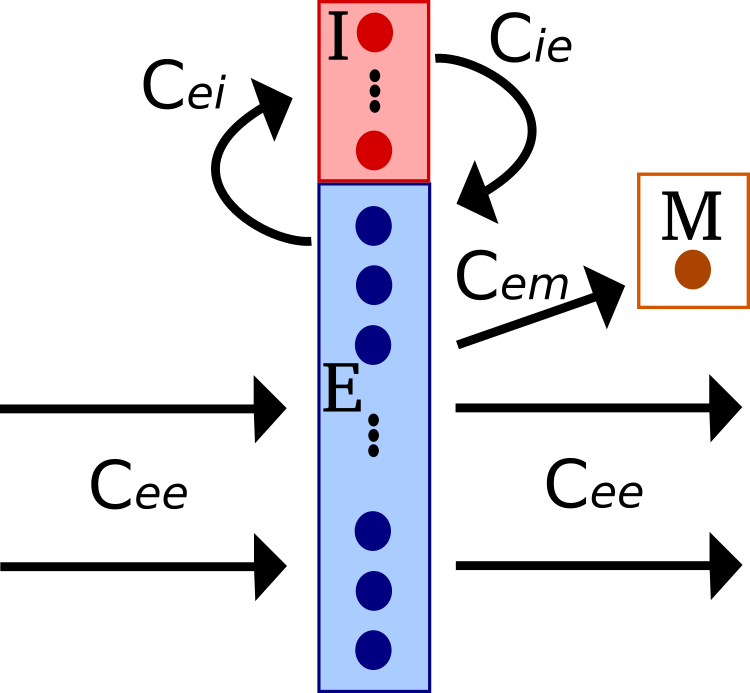}
      \caption{Connectivity of the population model. E = excitatory, I = inhibitory, M = monitor neuron. For $C_\mathrm{ei}$, $C_\mathrm{ie}$, $C_\mathrm{ee}$ and $C_\mathrm{em}$ see Table~\ref{table_connectivity}.}
      \label{PopulationConnectivity}
   \end{figure}

The neural network is a reservoir consisting of 300 populations of spiking neurons (unless specified otherwise), arranged in a three dimensional structure of 3x3 layers (Figure~\ref{overview}). Each excitatory neuron of a population connects to a neuron of another population with a probability proportional to the Euclidean distance between both populations (see Table~\ref{table_connectivity}). This distance-based connectivity is not only biologically plausible but also makes the simulation and the potential hardware implementation feasible as it reduces the overall number of connections. The delay of spike transmission between populations is fixed at 100ms. Each population consists of 40 neurons of the leaky-integrate-and-fire (LIF) type with exponentially decaying post-synaptic current (\textit{iaf\textunderscore psc\textunderscore exp}, as described in \cite{tsodyks2000synchrony}). Neuron parameters are close to the default, bioplausible values, or hand tuned for desired population response properties (Table~\ref{table_neuron_params}). The ratio of inhibitory/excitatory neurons is 1/4. Within a population, excitatory neurons connect to inhibitory neurons and vice versa (see Figure~\ref{PopulationConnectivity} and Table~\ref{table_connectivity}). All neurons in a population receive a white noise current of mean 0 and SD 2, this is important in maintaining a responsive population.\\

\subsection{The interface between neural network and body}

Interfacing the spiking neural network with the robot body requires translating spiking activity to analog values and vice versa. The motors of the actuated joints expect an analog value, the desired joint angle. Each motor has a readout neuron that provides this value. The parameters of the readout neurons have been adapted such that its membrane potential can be used directly as motor signal (see Table~\ref{table_readout_params}). Most importantly, the spiking threshold is set to infinity, preventing the neuron from firing which would reset the membrane potential. As a result the readout neuron is simply a leaky integrator of its incoming spikes. In the other direction, body sensor data is fed to the neural network. Therefore a DC current proportional to the values of a sensor is injected into a sensor population, whose activity then closely reflects the sensor stream. In this fashion the interface between the spiking network and the body is accomplished.\\

\begin{figure}[thpb]
      \centering
      \includegraphics[scale=0.3]{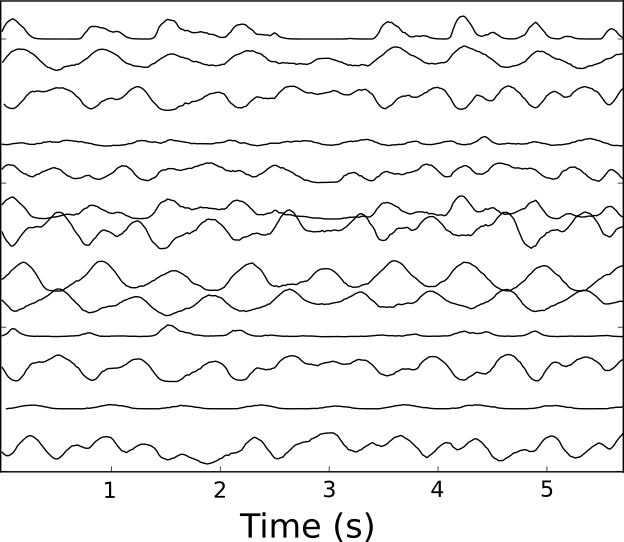}
      \caption{Membrane potentials of a few monitor neurons. Each monitor neuron represents the activity of one population of spiking neurons.}
      \label{example_monitorNeuron_activity}
   \end{figure}

The readout neurons are connected to all reservoir populations. The weights of these connections are learned with FORCE learning~\citep{sussillo2009generating}. Therefore, the reservoir states (i.e. the population activities) need to be known at all times. To observe the reservoir states, each population is monitored by a monitor neuron. Monitor neurons are identical to readout neurons, but are connected to a single reservoir population with unit weight. The membrane potential of the monitor neuron represents the population activity (Figure~\ref{example_monitorNeuron_activity} shows an example of the membrane potential of a few monitor neurons) and is used by the FORCE learning algorithm.\\

\begin{table}[h]
\begin{center}
\begin{tabular}{ c c}
\hline
Parameter & Value\\
\hline
Membrane resting potential [mV] & -65\\
Spiking threshold [mV] & -50\\
Post spike reset membrane potential [mV] & -75\\
Membrane capacitance [nF] & 0.2\\
Membrane time constant [ms] & 30\\
Duration refractory period [ms] & 2\\
Post-synaptic time constant [ms] & 0.5\\
\hline\\
\end{tabular}
\caption{Parameters of the LIF neuron model.}\label{table_neuron_params}
\end{center}
\end{table}

\begin{table}[h]
\begin{center}
\begin{tabular}{ c c}
\hline
Parameter & Value\\
\hline
Membrane resting potential [mV] & 0\\
Spiking threshold [mV] & $\infty$\\
Membrane capacitance [nF] & 0.2\\
Membrane time constant [ms] & 30\\
Post-synaptic time constant [ms] & 5.5\\
\hline\\
\end{tabular}
\caption{Parameters of the LIF readout and monitor neurons.}\label{table_readout_params}
\end{center}
\end{table}

\begin{table}[h]

\begin{center}
\begin{tabular}{c c c}
\hline
Connection & Variable & $P\mathrm{connect}$\\
\hline
Intra-population: Excitatory to inhibitory & $C_\mathrm{ei}$ & 0.1\\
Intra-population: Inhibitory to excitatory & $C_\mathrm{ie}$ & 0.1\\
Inter-population: Excitatory to excitatory & $C_\mathrm{ee}$ & $0.3e^{-D^{2}}$\\
Excitatory to monitor neuron & $C_\mathrm{em}$ & 1.0\\
\hline\\

\end{tabular}
\caption{Connectivity of the population model per connection type. Pconnect = connection probability between any two neurons. D = Euclidean distance between two populations.}
\label{table_connectivity}
\end{center}
\end{table}

\subsection{Gradual FORCE learning }\label{FORCEsection}

The aim of the learning is to find connection weights that make the membrane potential of the four readout neurons (see Figure \ref{overview}) produce the predefined target signals, i.e. the four motor signals as found by the CMA-ES optimization. \\

Reservoirs with feedback are challenging to train due to the feedback introducing delayed effects. FORCE learning allows to impose behaviour of a reservoir with feedback using the recursive least-squares algorithm. It suppresses unstable behaviour by reducing the error magnitude from the onset of the learning procedure.\\

By using monitor neurons, we effectively isolate the unweighted contribution of each reservoir population to a readout neuron. Since readout neurons and monitor neurons have identical parameters, the membrane potential of a readout neuron will be a linear combination of all monitor neuron membrane potentials. Therefore it is possible to use the monitor neurons membrane potentials as reservoir states, and FORCE learning can be applied in an identical fashion as with rate-based neural networks.\\

To make a stable closed loop system, the control signals are first learned in open loop with the control signals as input to the actuators. Subsequently, to ensure a smooth transition to closed loop control, the target signal and readout signal are gradually mixed (as described in~\cite{caluwaerts2013locomotion}). The contribution of the readout neuron is gradually increased during this transition. Finally, the system is capable of autonomously producing the target signals in a stable closed loop fashion (as detailed in the results section).\\

The regularization variable $\alpha$ of the FORCE learning algorithm must be selected large enough to prevent overfitting, but not too large as it could fail to approximate the target function sufficiently fast~\citep{sussillo2009generating}. After a parameter sweep, a value of 50 for $\alpha$ was observed to be effective. Furthermore, to ensure robustness of the closed loop system, it is necessary to insert sources of noise during learning. Here, impulse noise and gaussian noise were added to the sensor signals (see also Figure~\ref{overview}). Similarly, noise is added to the target signals. A low pass filter is added to smoothen the sensor signals before injecting them to the reservoir. The actuators also posses low-pass properties, which filters out some of the noise due to the implementation with spiking neurons.\\

\section{RESULTS}

\subsection{Gait Generation}

\begin{figure}[thpb]
      \centering
      \includegraphics[scale=0.4]{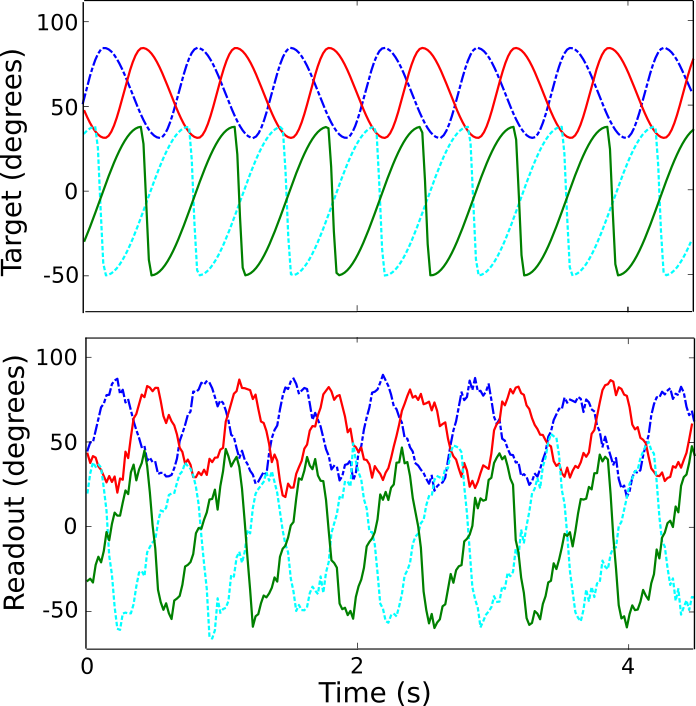}
      \caption{Walking Gait. Top pane: target motor signals (red, blue, green and cyan for front left, front right, hind left and hind right legs respectively). Bottom pane: readout signals during closed-loop control, after FORCE learning.}
      \label{GaitGeneration0}
   \end{figure}

\begin{figure}[thpb]
      \centering
      \includegraphics[scale=0.4]{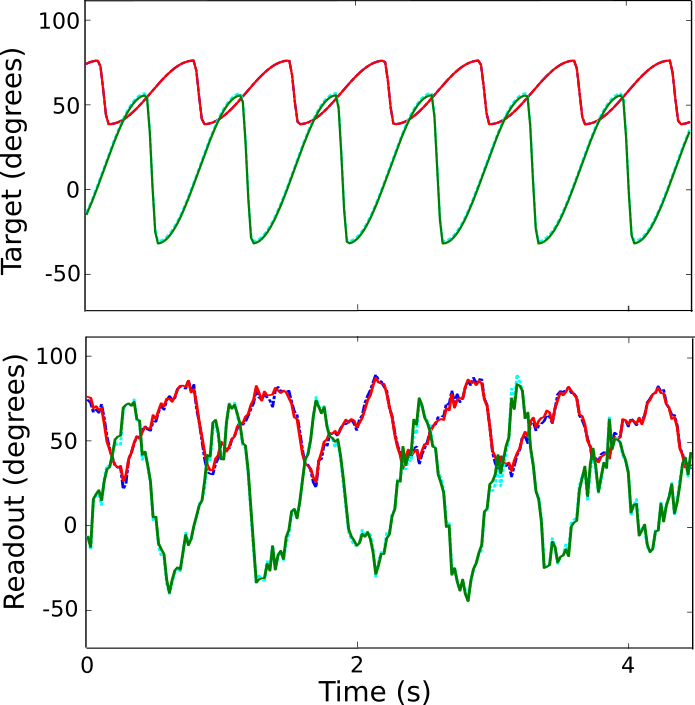}
      \caption{Bounding gait. Top pane: target motor signals (red, green for front, hind legs respectively) Bottom pane: readout signals during closed-loop control, after FORCE learning.}
      \label{GaitGeneration1}
   \end{figure}

The system is capable of learning and sustaining different gaits that have been found using CMA-ES. Figures~\ref{GaitGeneration0} and ~\ref{GaitGeneration1} display the target motor signals and the generated motor signals during a closed loop walking and bounding gait, respectively. In these experiments learning took 40 seconds simulated time for open loop training followed by 40 seconds for closed loop training. After learning, the gait generation is sufficiently robust to continue after disturbances such as moving the robot or stopping movement by turning the robot on its back for a while. The learned motor signal  are a bit noisy. This is mainly due to the rather small population sizes. \\

\subsection{Speed Control}

\begin{figure}[thpb]
      \centering
	  \includegraphics[scale=0.20]{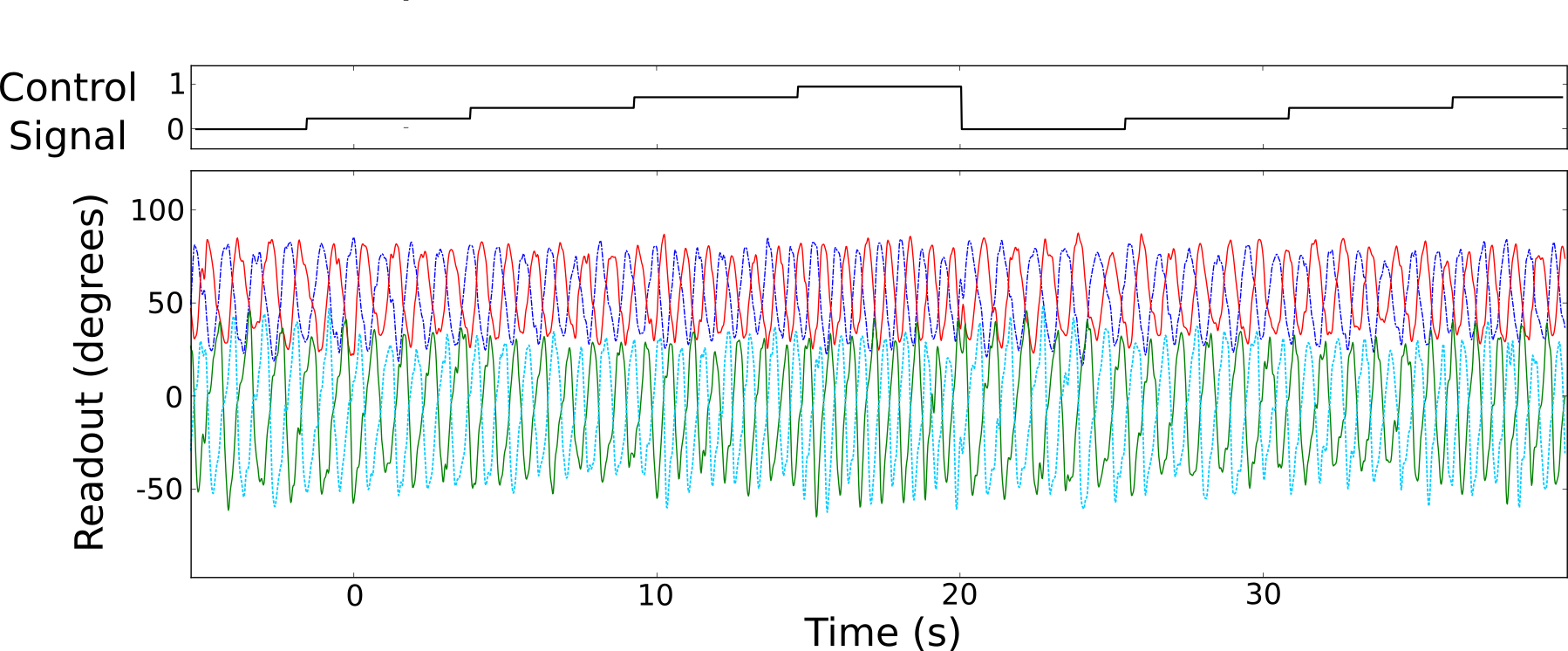}
      \caption{Tunable frequency post learning.}
      \label{multifreq}
   \end{figure}

In order to obtain gaits with tunable speed, we added an extra control input to the reservoir that serves as a control signal to control gait frequency. Similarly to sensor inputs, the control input is implemented as a DC current to the reservoir. During training, incremental frequencies of the same gait are paired with an incremental control signal. After learning, the control signal can be used to alter the frequency (Figure~\ref{multifreq}). The total learning time for this experiment was 200 seconds.\\

\subsection{Gait Transition}

\begin{figure}[thpb]
      \centering
      \includegraphics[scale=0.18]{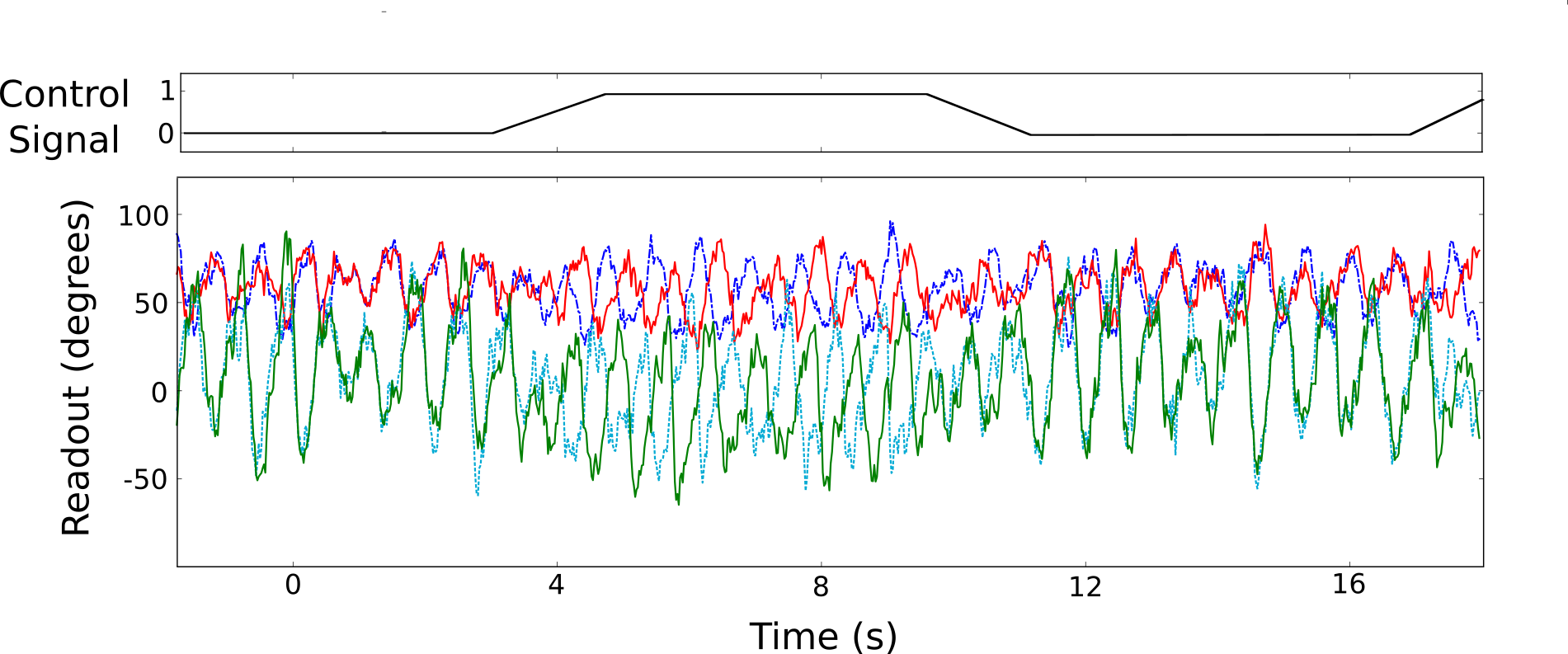}
      \caption{Gait transitioning controlled by external input.}
      \label{MultiGait_postFORCE}
   \end{figure}

Here, the network is trained to produce both gaits presented previously (walking and bounding). Again, a simple high level control input is used during and after training to control the gait transition (Figure~\ref{MultiGait_postFORCE}). For this experiment, the reservoir was made more powerful by increasing the number of populations to 600 and the number of neurons per population to 100. The total learning time for this experiment was 200 seconds. \\

\section{CONCLUSIONS}

Spiking neural networks could be advantageous for robotics. Potential benefits are energy efficient hardware implementations, efficient sensors and the possibility to apply learning principles observed in biological networks. This work proposes a novel architecture that enables the use of  populations of spiking neurons as reservoir units that complement and exploit the physical reservoir that the robot body is. Using only simple learning rules, stable closed loop locomotion control is achieved, even if only minimal sensor data is provided. As in biological spinal networks, the CPG output can be modulated by both simple high level inputs and body sensor input.\\

In this work, a population of spiking neurons is treated at the same conceptual level as a single artificial neuron, in order to use the same learning paradigm. However, a population of spiking neurons is potentially much more powerful due to the larger number of parameters, both the neuron and population parameters. In future work, the potential of the populations as unit for reservoir computing will be further investigated. Aditionally, an implementation on neuromorphic hardware (SpiNNaker) will allow to run the network in real time on the physical Tigrillo robot.\\

\section*{ACKNOWLEDGMENT}

This research was supported by the HBP Neurorobotics Platform funded from the European Union’s Horizon 2020 Framework Programme for Research and Innovation under the Specific Grant Agreement No. 785907 (Human Brain Project SGA2).





\bibliographystyle{model5-names}
\biboptions{authoryear}
\bibliography{bibliography.bib}







\end{document}